\pdfoutput=1
\documentclass[11pt]{article}
\usepackage[final]{acl}
\usepackage{times}
\usepackage{latexsym}
\usepackage[T1]{fontenc}
\usepackage[utf8]{inputenc}
\usepackage{microtype}
\usepackage{inconsolata}
\usepackage{graphicx}
\usepackage{booktabs}
\usepackage{amsmath}
\usepackage{amssymb}
\usepackage{enumitem}
\usepackage{tikz}
\usepackage{algorithm}
\usepackage{algpseudocode}
\usepackage{subcaption}
\usepackage{multirow}
\usepackage{soul}
\usepackage{framed}
\usepackage{xcolor}
\usepackage[most]{tcolorbox}
\usepackage{amsmath,amssymb,amsthm}
\newtheorem{proposition}{Proposition}

\usetikzlibrary{positioning, arrows.meta, calc, decorations.pathreplacing}
\usepackage{geometry}

\usetikzlibrary{arrows.meta,positioning,shapes.geometric,fit,backgrounds,calc,decorations.markings}

\usepackage{helvet}

\definecolor{kbpurplelight}{HTML}{EDEBF8}
\definecolor{amberlight}{HTML}{F8EFDC}
\definecolor{annotgreen}{HTML}{0F6E56}
\definecolor{annotgreenlight}{HTML}{E3F1EB}
\definecolor{relationred}{HTML}{993556}
\definecolor{kbblue}{HTML}{185FA5}
\definecolor{missred}{HTML}{A32D2D}
\definecolor{missredlight}{HTML}{F8E5E5}
\definecolor{inputgray}{HTML}{E8E7E1}
\definecolor{inputborder}{HTML}{8E8D86}
\definecolor{arrowgray}{HTML}{6F6F6F}

\usepackage{tabularx}
\usepackage{array}

\newcolumntype{L}[1]{>{\raggedright\arraybackslash}p{#1}}
\newcolumntype{Y}{>{\raggedright\arraybackslash}X}

\title{KARLA: Knowledge-base Augmented Retrieval for Language Models}

\author{François Crespin, \quad Fabian M. Suchanek, \quad Nils Holzenberger \\
  Télécom Paris \\
  Institut Polytechnique de Paris, France \\
  \texttt{\{francois.crespin, fabian.suchanek, nils.holzenberger\}@telecom-paris.fr}}

\begin{document}
\maketitle

\begin{abstract}
We propose a new method that allows an LLM to automatically pull in factual knowledge from a knowledge base during token generation. This means that (1)~factual knowledge in the LLM output can be updated without retraining the LLM, (2)~facts in the LLM output can be traced to the knowledge base for transparency and explainability, and (3)~smaller models can achieve the same factual accuracy as larger models. Our core idea is to train the model to produce special tokens that trigger a query to the knowledge base. Our experiments show that our method improves factual grounding in both short and long-form generation, and allows factual revisions to take effect through KB edits rather than parameter updates.\footnote{Code and data are included in the supplementary material and will be made available publicly.}

\end{abstract}

\section{Introduction}

Language models can acquire factual knowledge during pre-training, enabling them to answer factual questions without explicit access to external resources \citep{petroni2019language}. However, this knowledge is stored implicitly in model parameters, making individual facts difficult to inspect, correct, or selectively update. This creates a fundamental limitation for factual generation: models may hallucinate \citep{maynez2020faithfulness}, become temporally misaligned as world knowledge changes \citep{lazaridou2021mindgapassessingtemporal, luu2022time}, or over-rely on memorized facts even when external evidence provides a conflicting update \citep{longpre2021entity}. Furthermore, their factual recall is uneven across relations and entities, and degrades for long-tail entities \citep{mallen2023not}. Finally, relying on parameters to store factual knowledge does not scale efficiently: factual knowledge capacity increases with model size, and conversely, memorizing large-scale knowledge bases may require extremely large models and training budgets \citep{lu2024scaling}.

\begin{figure}[t]
  \centering
  \resizebox{\columnwidth}{!}{%
    \begin{tikzpicture}[
    font=\small,
    rel/.style={
      rectangle, rounded corners=1.5pt,
      draw=purple!70!black, fill=purple!15, text=purple!40!black,
      font=\footnotesize\ttfamily, inner sep=2pt, minimum height=4mm
    },
    subj/.style={
      rectangle, rounded corners=1.5pt,
      draw=teal!70!black, fill=teal!18, text=teal!35!black,
      font=\footnotesize\bfseries, inner sep=2pt, minimum height=4mm
    },
    obj/.style={
      rectangle, rounded corners=1.5pt,
      draw=orange!80!black, fill=orange!25, text=orange!45!black,
      font=\footnotesize\bfseries, inner sep=2pt, minimum height=4mm
    },
    plain/.style={font=\footnotesize, inner sep=1pt},
    qline/.style={semithick, gray!55!black},
    qarrow/.style={->, >=Stealth, semithick, gray!50!black},
    rarrow/.style={->, >=Stealth, semithick, orange!70!black},
    llm/.style={
      rectangle, rounded corners=3pt,
      draw=blue!65!black, fill=blue!10, text=blue!35!black,
      font=\footnotesize\bfseries, inner sep=4pt, minimum height=6mm
    },
    garrow/.style={->, >=Stealth, semithick, blue!55!black},
  ]
  \node[subj] (wsubj) {Paris};
  \node[plain, right=1mm of wsubj] (w2)   {has a population of};
  \node[rel,   right=1mm of w2]    (wrel) {$\langle$populationTotal$\rangle$};
  \node[obj,   right=1mm of wrel]  (wobj) {2{,}047{,}602};

  \draw[decorate, decoration={brace, amplitude=5pt, raise=3pt},
        semithick, blue!60!black]
    (wsubj.north west) -- (wrel.north east);
  \coordinate (bracemid) at ($(wsubj.north west)!0.5!(wrel.north east) + (0, 0.32)$);

  \node[llm] at ($(bracemid) + (0, 0.85)$) (karla) {KARLA Model};

  \draw[garrow] (karla.south) -- (bracemid);

  \coordinate (junction) at ($(wsubj.south)!0.5!(wrel.south) + (0, -0.70)$);
  \coordinate (kbcenter) at ($(junction) + (0, -0.90)$);
  \begin{scope}[shift={(kbcenter)}]
    \filldraw[fill=orange!10, draw=orange!75!black, semithick]
      (-0.50, 0) -- (-0.50, -0.65)
      arc[start angle=180, end angle=360, x radius=0.50, y radius=0.14]
      -- (0.50, 0)
      arc[start angle=0, end angle=180, x radius=0.50, y radius=0.14];
    \filldraw[fill=orange!20, draw=orange!75!black, semithick]
      (0, 0) ellipse[x radius=0.50, y radius=0.14];
    \draw[orange!55!black, dashed, thin]
      (-0.50, 0) arc[start angle=180, end angle=0, x radius=0.50, y radius=0.14];
    \node[font=\footnotesize\bfseries, text=orange!45!black] at (0, -0.40) {KB};
  \end{scope}
  \coordinate (kbTop)   at ($(kbcenter) + (0,    0.14)$);
  \coordinate (kbRight) at ($(kbcenter) + (0.50, -0.325)$);
  \draw[qline]  (wsubj.south) -- (junction);
  \draw[qline]  (wrel.south)  -- (junction);
  \draw[qarrow] (junction)    -- (kbTop);
  \node[font=\scriptsize\itshape, text=gray!40!black, anchor=east, inner sep=2pt]
    at ($(junction)!0.5!(kbTop) + (-0.08, 0)$) {Query KB};
  \draw[rarrow, rounded corners=3pt]
    (kbRight) -- (wobj.south |- kbRight) -- (wobj.south);
  \node[font=\scriptsize\itshape, text=orange!55!black, anchor=north, inner sep=2pt]
    at ($(kbRight)!0.5!(wobj.south |- kbRight) + (0, -0.03)$) {Returns KB value};
\end{tikzpicture}
  }
  \caption{KARLA in action. 
  }
  \label{fig:tokens}
  \vspace{-\baselineskip}
\end{figure}
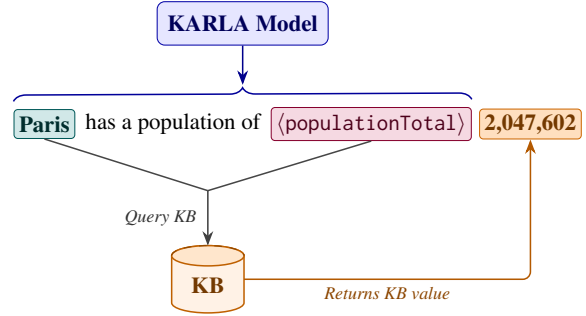

Retrieval-augmented generation (RAG) methods \citep{lewis2020retrieval,guu2020retrieval,borgeaud2022improving} and tool-use approaches \citep{yao2022react,schick2023toolformer} allow pulling in factual information from external textual resources or tools. 
These methods improve factual grounding and make some knowledge updates possible through changes to the retrieval corpus or tool backend. However, retrieval alone does not eliminate parametric factual reliance. The model may ignore retrieved evidence, blend it with memorized knowledge, or generate claims that are not attributable to the retrieved source \citep{ravi2024lynx,NYTimes2023,llm-search-engine}. 

In this paper, we propose KARLA, a method that entices a language model to query factual knowledge from a knowledge base (KB, also called knowledge graph), i.e., a structured repository of subject-relation-object triples. 

This endeavor is challenging for several reasons: First, the model must issue correct queries, i.e., it must identify the subject and relation of a query that has a chance to deliver a result. Second, the model must learn to issue correct queries also for entities that were not seen during training, and about which it has no parametric knowledge. (While it appears easy to determine that population size is a reasonable query for ``Paris'', what is a reasonable query for ``9-hydroxyrisperidone''?)
Finally, our method has to work without any manual supervision or intervention.

Our key idea is to generate a synthetic corpus directly from the KB, interleaved with relation-specific query tokens.
The model is then fine-tuned on this corpus to issue these tokens whenever it is about to generate factual information (Figure~\ref{fig:tokens}). During inference, we replace these tokens with the results of a query to the KB.
Our goal is to separate linguistic competence from factual knowledge: the language model handles interpretation and generation, while the KB provides the atomic factual values. This brings several advantages: 
\begin{enumerate}[nolistsep,noitemsep,leftmargin=4mm]
    \item Factual knowledge can be \textbf{updated} in the KB at virtually no cost, without retraining the model. Any change is effective immediately. 
    \item Knowledge that was pulled in from the KB can be marked as such, thus leaving a \textbf{provenance} that serves transparency and explainability. Every KB-sourced span is verifiable against a specific fact in the KB, a guarantee that neither vanilla generation nor RAG can deliver. 
    \item The model can be much smaller, as all factual knowledge resides in the KB and not in the parameters. KARLA models \textbf{scale} not with the number of facts (as parametric knowledge does), but with the number of relations (which is typically small), meaning that our approach allows even small models to achieve high factual accuracy with KBs of near-arbitrary size.
\end{enumerate}

\noindent Indeed, our experiments on factual question answering, long-form generation, and counterfactual KB-update settings show that (1)~the model will generate updated information from the KB even when this contradicts its own parametric outdated knowledge, (2)~a substantial part of generated long-form text can be sourced to the KB, and (3)~small models such as Qwen0.6B comfortably beat larger models in factual accuracy when enhanced with our method.

\section{Related Work}

\paragraph{Language models as KBs.}
Pretrained models store facts implicitly \citep{petroni2019language}, but their capacity is limited to $\approx$2 bits/parameter \citep{allen2023physics,allen2025physics} and uneven for long-tail entities \citep{mallen2023not}. Theoretical work suggests that external lookup outperforms memorization under these constraints \citep{houliston2025provable}, which motivates our shift of factual storage outside the model.

\paragraph{Retrieval-augmented generation}$\!\!\!\!$(RAG)
 places passages retrieved from external documents in the context window of the model \citep{lewis2020retrieval, guu2020retrieval, borgeaud2022improving}. However, the model can choose to ignore this information, or hallucinate information that was not given \citep{ravi2024lynx,NYTimes2023,llm-search-engine}, as we also show in our experiments.
 
\paragraph{Tool-use approaches}$\!\!\!\!$let the model issue calls to external systems \citep{schick2023toolformer,yao2022react, wang2024toolgen}, sometimes via learned special tokens per tool \citep{hao2023toolkengpt}, but rely on annotated demonstrations of tool use that do not scale to the relation inventory of a typical KB. KARLA instead generates training data synthetically from the KB itself.
KBLaM \citep{wang2024kblam} integrates a knowledge base into a language model through continuous key-value representations that support dynamic updates without retraining. However, this method holds the entire KB via an attention mechanism and attends over all knowledge tokens at query time. Hence, it is applicable only to small KBs. KARLA, in contrast, is designed for large KBs such as YAGO \cite{suchanek2007yago} with millions of triples.

\paragraph{Large Memory Language Models}$\!\!\!\!$(LMLMs) \citep{zhao2025pre} are pre-trained from scratch with retrieved factual values masked from the loss, so that the model learns lookup behavior rather than memorization. The method takes as input a corpus of text, and builds up the KB from there. Our method, in contrast, is designed for the scenario where the user already possesses a KB, and wants the language model to answer with factual information from this source.

\section{Methodology: KARLA}\label{sec:karla}

\subsection{Problem Setup}

\paragraph{Approach.} We are given a pre-trained language model and a knowledge base (KB). 
Formally, a KB is a set of triples of the form $\langle{}s, r, o\rangle{}$, where $s$ is a subject entity, $r$ is a relation (or predicate) from a closed set $\mathcal{R}$, and $o$ is the corresponding object, as in $\langle$Paris, populationTotal, 2,047,602$\rangle$. Our objective is to fine-tune the model so that it interleaves natural-language generation with inline queries of the form $\langle{}s, r, ?\rangle{}$. These inline queries are then replaced by an object $o$ for which $\langle{}s, r, o\rangle{}$ is in the KB (see Figure~\ref{fig:tokens}). 

\paragraph{Inline queries.} In practice, the inline query $\langle{}s, r, ?\rangle{}$ with result $o$ is expressed as
\begin{equation*}
  \langle{}r\rangle \, \langle\texttt{subj}\rangle{}s\langle/\texttt{subj}\rangle \, \langle\texttt{KB}\rangle{}o : o_{\text{desc}} \langle/\texttt{KB}\rangle
\end{equation*}
In this sequence, $\langle{}r\rangle$
is a relation-specific trigger token. 
The tag $\langle\texttt{subj}\rangle$ marks the subject, which is used to query the KB. The model will be trained to identify the subject here from the preceding part of the sentence. The tag $\langle\texttt{KB}\rangle$ marks the answer to the query that was retrieved from the KB. It is included in the output sequence of tokens as soon as the inline query is generated, so as to allow the model to condition the following tokens on $o$. The text $o_{\text{desc}}$ is a short description of the object as retrieved from the KB (e.g., by help of the relation \texttt{schema:description}). It provides information about $o$ that the model may not have in its parameters, in particular for long-tail or unseen entities.
If the query has several answers in the KB, one of them is picked at random. This is motivated by the use case of long-form generation, where a single object will fulfill the expected role (as in ``Elvis Presley won the $\langle$award$\rangle\langle\texttt{subj}\rangle$Elvis Presley$\langle/\texttt{subj}\rangle\langle\texttt{KB}\rangle$Grammy Award$ \langle/\texttt{KB}\rangle$'', where the Grammy Award is one of the awards of Elvis).

\paragraph{Predicate Representation and Initialization.} We represent each relation as an atomic special token $\langle{}r\rangle$ rather than as a plain-text label. This has the advantage that the model cannot generate relations that do not exist in the KB.
It also reduces per-call inference cost, because each retrieval action emits one token rather than the several sub-tokens of a plain-text predicate label. 
To speed up convergence, we initialize the new embeddings from the base tokenizer.  Let $E \in \mathbb{R}^{|\mathcal{V}| \times d}$ denote the embedding matrix of the original tokenizer where $\mathcal{V}$ is the vocabulary set and $d$ is the embedding dimension. For a relation $r$,
we define the embedding of the corresponding predicate token $E_{\langle r \rangle}$ as the mean of its constituent sub-token embeddings \citep{wang2024toolgen}.
The added vocabulary introduces $|\mathcal{R}| \cdot d$ embedding parameters, a negligible fraction of the base model size. We provide an ablation against a plain-text predicate representation in Appendix~\ref{app:predicate-ablation}.

\subsection{Generating the training corpus}
\label{sec:synth}

To train the model to interact with the KB, we require a corpus in which natural language is interleaved with KB queries. One could, of course, take an existing corpus of text, and weave in the KB queries. However, this  
would require jointly solving several information extraction tasks, including named entity recognition \citep{zaratiana2024gliner}, entity linking and entity disambiguation \citep{wu2020scalable, haffoudhi2026lela}, and relation extraction \citep{cabot2021rebel}. In addition, naturally occurring text provides highly imbalanced coverage of KB relations, with long-tail predicates appearing too rarely to support robust learning \citep{josifoski2023exploiting}. We thus generate the training corpus synthetically from the KB.

For this purpose, we first have to sample facts from the KB. A naive uniform sampling would generate a corpus of disconnected facts, in which each sentence expresses a single fact and the phrasings in which the facts appear are severely limited. Therefore, we generate texts that talk about several facts of a single entity before moving on to the next entity. If we sample entities instead of facts, though, we cannot ensure that all relations appear equally often.
We therefore formulate the sampling as an iterative greedy process designed to satisfy a target distribution across the relational schema. Let $\mathcal{R}$ be the set of KB predicates and $T$ the target frequency for each relation. At each iteration $i$, we identify the least used relation $r^* \in \mathcal{R}$. 
We then sample an entity $e$ uniformly at random from the set $\{e \mid \exists (e, r^*, o) \in \text{KB}\}$. For the selected entity $e$, we retrieve its fact neighborhood $\mathcal{F}(e)=\{\langle{}e, r, o\rangle \mid \langle{}e, r, o\rangle \in \text{KB}\}$, i.e., the set of all KB triples with $e$ as subject. When a relation $r$ admits several objects, we sample one uniformly at random. 
We remove from the neighborhood all facts with relations $r$ that have already appeared $2\times{}T$ times in previous samples. We then sample $k$ facts from this reduced neighborhood.
The complete sampling algorithm 
is provided in Appendix~\ref{app:sampling_algorithm}. We also prove in Proposition \ref{prop:balance} in the same appendix that our method yields per-relation counts bounded between $T$ and $2\times{}T$ regardless of the KB's natural relation distribution (as we sample with replacement). 

Once the facts are sampled, we prompt a teacher LLM (GPT-5 mini \citep{openai2025gpt5developers} in our experiments) to produce the training corpus. For this purpose, we iteratively feed it with each sampled entity subgraph, together with natural-language descriptions of the relations, and prompt it to generate an encyclopedia-style paragraph (approximately 250 tokens) with inline queries. Prompts are provided in Tables~\ref{tab:verbalizer_prompt_yago} and~\ref{tab:verbalizer_prompt_primekg} (Appendix \ref{app:prompts}).
We validate each generated example using two checks: (i)~for every annotated object mention, the text span 
must match exactly the referenced object $o$; and (ii)~every sampled relation in $\mathcal{F}(e)$ must appear in the output. If either check fails, the teacher receives structured feedback describing missing relations and parsing errors, and generation is retried up to two times, after which the subgraph is discarded. This affects less than $0.5\%$ of subgraphs in our experiments.

\subsection{Training}

We then train our input model on the generated corpus. 
Let $\tilde{x} = (\tilde{x}_1, \ldots, \tilde{x}_M)$ 
denote a sequence of tokens with inline queries and their KB-returned objects. 
Since training covers only a small subset of the KB, the model must learn to handle cases where the KB contains no matching fact. In 10\% of inline queries (sampled independently), we replace the successful lookup result with the failure token $\langle\texttt{KB\_FAIL}\rangle$. This teaches the model to recover from empty or unresolved KB queries by falling back to its parametric knowledge.

To enforce the separation of natural language and KB knowledge, we employ a masked next-token prediction loss. The loss is defined as: 
\begin{equation*}
  \mathcal{L}(\theta) = -\sum_{t=1}^{N} m_t \cdot \log p_\theta\left(\tilde{x}_t \mid \tilde{x}_{<t}\right)
\end{equation*}
The binary mask $m_t \in \{0, 1\}$ is constructed to gate the gradient flow:
\begin{equation*}
  m_t = 
  \begin{cases} 
    0 & \text{if } \tilde{x}_t \in \text{span}(\langle\texttt{KB}\rangle, \dots, \langle/\texttt{KB}\rangle) \\
    1 & \text{otherwise}
  \end{cases}
\end{equation*}
By setting $m_t = 0$ for tokens within the retrieval-result span, the model is not penalized for failing to predict the returned object $o$ from the context alone. Instead, supervision is concentrated on the surrounding natural language and on the retrieval query that obtains the relevant KB evidence. This reduces direct supervision pressure to store KB facts in the model parameters and encourages the model to condition factual generation on the retrieved KB object.

\subsection{Inference}\label{ref:inference}

Before generation begins, we pre-process the prompt to bring any entities into the form that KARLA-models can digest. More precisely, any entity mention $e$ in the prompt is replaced by \mbox{⟨KB⟩$e^*$: $e^*_{desc}$⟨/KB⟩}, where $e^*$ is the canonical KB entity of $e$. To determine $e^*$, we use a two-stage pipeline inspired by \citet{haffoudhi2026lela}:
\begin{enumerate}[nolistsep,noitemsep,leftmargin=4mm]
    \item \textbf{Candidate Retrieval:} A bi-encoder (all-MiniLM-L6-v2 \citep{sentence_transformers_all_minilm_l6_v2}) retrieves the top 10 candidate entities from a FAISS \citep{douze2025faiss} index built over entity embeddings.
    \item \textbf{Contextual Re-ranking:} A small language model (Qwen3 1.7B) then re-ranks these candidates using the surrounding generation context to resolve ambiguities and determine $e^*$.
\end{enumerate}
This allows the model to condition any following tokens on $e^*$. 

The generation then proceeds auto-regressively.
A KB query is triggered when the generation terminates in a sequence of the form:
\begin{equation*}
    \langle{}r\rangle\, \langle\texttt{subj}\rangle \, s \, \langle/\texttt{subj}\rangle 
\end{equation*}
where $r$ is a relation-specific token. 
When detected, the model generation is paused to resolve $s$ to a canonical entity $s^*$ of the KB, in the same way as before. The system then attempts to fetch an object $o$ such that $\langle{}s^*, r, o\rangle \in \text{KB}$. The generation continues by appending either the retrieved object or a failure signal to the sequence:
\begin{equation*}
  \text{Output} = 
  \begin{cases} 
    \langle\texttt{KB}\rangle \, o : o_{\text{desc}} \, \langle/\texttt{KB}\rangle & \text{if } \langle{}s^*, r, o\rangle \in \text{KB} \\
    \langle\texttt{KB\_FAIL}\rangle & \text{otherwise}
  \end{cases}
\end{equation*}
In the event of a $\langle\texttt{KB\_FAIL}\rangle$, the model falls back to its internal parametric weights to complete the sequence. This makes it possible to distinguish content that comes from the KB from content that comes from parametric memory.

\section{Experiments}

\subsection{Experimental setup}

\paragraph{Knowledge bases.}
We instantiate KARLA on two knowledge bases with complementary domains:
YAGO~\citep{suchanek2024yago}, a general-domain encyclopedic KB, and
PrimeKG~\citep{chandak2023building}, a biomedical KB.
Using both resources allows us to evaluate whether the method generalizes across open-domain and specialized scientific knowledge.
Detailed statistics for the KBs are in Appendix~\ref{app:kb_details}.

\begin{figure}[t]
    \centering
    \includegraphics[width=\columnwidth]{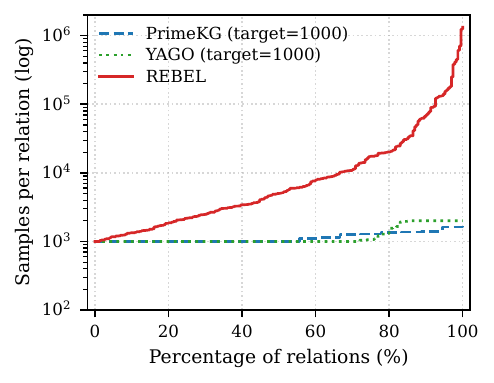}
    \caption{
Per-relation sample counts, sorted in ascending order. For each percentile of relations on the x-axis, the y-axis (log scale) shows the number of available samples. 
}
    \label{fig:balanced_sampling-figure}
    \vspace{-\baselineskip}

\end{figure}

\paragraph{Training data.}
For both KBs, we construct the training corpus as explained in Section~\ref{sec:synth}. We sample $T = 1000$ triples per relation with $k = 7$ relations per subject, over $|\mathcal{R}| = 99$ relations for YAGO (after filtering non-semantic predicates such as \texttt{yago:url}) and $|\mathcal{R}| = 18$ for PrimeKG. As shown in Figure~\ref{fig:balanced_sampling-figure}, this yields a near-balanced distribution of relations in the corpus.
The resulting training corpus contains $36{,}929$ paragraphs for YAGO and $10{,}470$ paragraphs for PrimeKG (example outputs are given in Tables~\ref{tab:verbalizer_prompt_yago} and~\ref{tab:verbalizer_prompt_primekg} in Appendix \ref{app:prompts}).
We also generate a test corpus with the same procedure, with subjects different from the ones in the training corpus.
This yields $1{,}389$ YAGO and $382$ PrimeKG paragraphs.

\paragraph{KARLA configurations.} All KARLA-models are based on
Qwen 3-Base \citep{yang2025qwen3} and fine-tuned with LoRA
\citep{hu2022lora} applied to all attention and MLP projection
matrices, following \citet{schulman2025lora}. We keep the
optimization setup fixed across model sizes: learning rate
$10^{-5}$, LoRA rank $64$, and scaling factor $\alpha{=}128$
(higher rank and $\alpha$ improve convergence, see
Fig.~\ref{fig:tool-learning-accuracy} Appendix \ref{app:LoRA-scaling}). We train on YAGO for 3
epochs and on PrimeKG for 5 epochs, holding out 10\% of each
corpus for validation; full training details are in
Appendix~\ref{app:Training-details}.
We evaluate under several configurations that share the same KB, and the same training corpus, differing only at inference:
\begin{description}[leftmargin=5mm,nolistsep,noitemsep]
  \item[KARLA:] standard configuration (Section~\ref{sec:karla}).
  \item[KARLA-no-desc:] same as KARLA, removing the entity description $e^*_{desc}$ from the prompt. 
  \item[KARLA-empty-KB:] same as KARLA, but every query returns $\langle \text{KB\_FAIL} \rangle$.
  \item[KARLA-raw:] same model, evaluated on raw passages without inline queries.
\end{description}
 
\paragraph{Competitors.}
We compare against four families of competitors:
\begin{description}[leftmargin=5mm,nolistsep,noitemsep]
    \item[Base LM:] Qwen-3 with no fine-tuning, probing parametric recall.
    \item[Parametric SFT:] LoRA fine-tuning on the same synthetic corpus but \emph{without} tool markup, forcing facts into parameters.
    \item[1-hop graph RAG:] the model is given the 1-hop neighborhood of the target entity in the KB, retrieved by entity linking, in its context window (details in Appendix~\ref{par:baseline_1hop_rag}).
    \item[Tool-schema guided:] prompted (not fine-tuned) tool use, where the relation schema is provided in the system prompt (details in Appendix~\ref{par:baseline_tool_schema}).
    \item[LMLM:] the closest published baseline, which trains an externalized-memory LM \emph{from scratch} \citep{zhao2025pre}.
\end{description}

\paragraph{Experiments.}
First, we study the impact of KARLA on the perplexity of the generated tokens (Section~\ref{sec:perplexity}). Second, we study the impact of KARLA on the factuality of the generated text (Section~\ref{sec:exp-factuality}). Finally, we study how KARLA fares when the KB contradicts the LLM's parametric knowledge (Section~\ref{sec:exp-update}).

\subsection{KARLA reduces perplexity}\label{sec:perplexity}

To analyze how KARLA benefits from executing queries over the KB, 
we measure the perplexity of KARLA on our test corpus. 
We ask whether the reduction in uncertainty from conditioning on KB evidence is large enough to compensate for the cost of generating the query.
For the competitors, we report standard perplexity:
\begin{equation*}
\mathrm{PPL}(x)
=
\exp\left(
-\frac{1}{N}
\sum_{i=1}^{N}
\log p_\theta(x_i \mid x_{<i})
\right)
\end{equation*}
For KARLA models, we report target-normalized masked perplexity.
Let $m_j \in \{0,1\}$ denote a scoring mask over the augmented sequence.
The mask is set to zero for KB-returned objects, since these tokens are supplied by the external executor rather than predicted by the model.
It is set to one for all tokens generated by the model, including test tokens and query tokens.
We define:
\begin{equation*}
\mathrm{PPL}_{\mathrm{aug}}(\tilde{x})
=
\exp\left(
-\frac{1}{N}
\sum_{j=1}^{M}
m_j
\log p_\theta(\tilde{x}_j \mid \tilde{x}_{<j})
\right)
\end{equation*}
\noindent This is not the ordinary perplexity of the augmented sequence: the denominator is the number of tokens $N$ in the non-augmented sequence, not the number of scored tokens nor the full augmented length $M$.
The metric thus measures the description length of the original passage while charging the model for the additional inline query tokens.
A lower score means that the evidence obtained through KB execution reduces uncertainty enough to outweigh the overhead of producing the KB query. 

\begin{table}[t]
\centering
\small
\setlength{\tabcolsep}{5pt}
\begin{tabular}{llrr}
\toprule
\textbf{Model} & \textbf{Setup} & \textbf{YAGO} & \textbf{PrimeKG} \\
\midrule
\multirow{4}{*}{Qwen 0.6B}
  & KARLA              & \textbf{7.09}  & \textbf{3.96} \\
  & KARLA-empty-KB     & 9.27           & 5.61 \\
  & KARLA-raw          & 11.17          & 6.38 \\
  & Raw-text SFT       & 8.79           & 5.08 \\
\midrule
\multirow{4}{*}{Qwen 1.7B}
  & KARLA              & \textbf{6.05}  & \textbf{3.36} \\
  & KARLA-empty-KB     & 7.65           & 4.59 \\
  & KARLA-raw          & 9.16           & 5.11 \\
  & Raw-text SFT       & 7.28           & 4.28 \\
\midrule
\multirow{4}{*}{Qwen 4B}
  & KARLA              & \textbf{5.32}  & \textbf{2.96} \\
  & KARLA-empty-KB     & 6.54           & 3.89 \\
  & KARLA-raw          & 7.61           & 4.47 \\
  & Raw-text SFT       & 6.27           & 3.75 \\
\midrule
\multirow{4}{*}{Qwen 8B}
  & KARLA              & \textbf{5.08}  & \textbf{2.84} \\
  & KARLA-empty-KB     & 6.13           & 3.75 \\
  & KARLA-raw          & 7.06           & 4.20 \\
  & Raw-text SFT       & 5.77           & 3.51 \\
\bottomrule
\end{tabular}
\caption{
Perplexity on the synthetic held-out set. KARLA and KARLA-empty-KB report normalized perplexity. Other rows report standard perplexity.
}
\label{tab:perplexity}
\vspace{-\baselineskip}
\end{table}
Table~\ref{tab:perplexity} shows that KARLA improves this target-normalized score across both KBs and all model sizes.
Compared with the Parametric SFT baseline, KARLA models lower perplexity for every evaluated Qwen~3 model.
The mean score over the four model sizes decreases from $7.03$ to $5.89$ on YAGO and from $4.15$ to $3.28$ on PrimeKG, corresponding to relative reductions of $16.2\%$ and $21.0\%$.
These gains hold despite the fact that KARLA models are penalized for generating queries, indicating that learned KB access provides useful evidence beyond what is captured by parametric-only fine-tuning. In the evaluation corpus, $22\%$ of YAGO tokens and $13\%$ of PrimeKG tokens are supplied directly by KB retrieval. A comparable share of the generated text is grounded in an explicit triple.

KARLA-empty-KB degrades performance from $5.89$ to $7.40$ on YAGO and from $3.28$ to $4.46$ on PrimeKG on average, showing that the improvement is not explained merely by tool syntax, sequence formatting, or tokenization artifacts, but by the actual values returned by the KB queries. KARLA-raw falls below Parametric SFT, indicating that KARLA trades standard text performance for inline-query specialization.

\begin{table}[t]
\centering
\small
\setlength{\tabcolsep}{4pt}
\resizebox{\columnwidth}{!}{%
\begin{tabular}{ll rrr rrr}
\toprule
& & \multicolumn{3}{c}{\textbf{YAGO}} & \multicolumn{3}{c}{\textbf{PrimeKG}} \\
\cmidrule(lr){3-5} \cmidrule(lr){6-8}
\textbf{Setup} & \textbf{Model} & Subj. & Rel. & Both & Subj. & Rel. & Both \\
\midrule
KARLA & Qwen 0.6B & 99.4 & 87.2 & 86.7 & 96.9 & 94.7 & 91.8 \\
KARLA & Qwen 1.7B & 99.6 & 89.9 & 89.5 & 98.1 & 96.6 & 94.9 \\
KARLA & Qwen 4B   & 99.7 & 91.2 & 91.0 & 99.2 & 97.8 & 97.0 \\
KARLA & Qwen 8B   & 99.7 & 88.1 & 87.9 & 99.4 & 95.3 & 94.8 \\
\bottomrule
\end{tabular}%
}
\caption{
Inline-query exact-match accuracy ($\%$) on the test corpus, broken down by predicted subject (Subj.), predicted relation (Rel.), and joint prediction (Both requiring the subject and relation to be correct).
}
\label{tab:retrieval-path-diagnostics}
\vspace{-\baselineskip}
\end{table}

We next assess inline-query accuracy on the test corpus, measured separately for the relation token, the subject span, and their joint prediction.
Table~\ref{tab:retrieval-path-diagnostics} shows that KARLA can indeed predict the subject and the relation with high accuracy. The accuracy is slightly lower for the relations on YAGO,
likely because its relations are more heterogeneous and because multiple predicates can be plausible from similar textual contexts. The relation accuracy on PrimeKG reaches $97.8\%$, indicating that its biomedical schema provides a more learnable relation-selection problem.

Together with the empty-KB ablation, this diagnostic supports the interpretation that KARLA's gains come from learned use of external evidence rather than from sequence-format artifacts. The model reliably learns to ground subjects and to issue executable queries.

\subsection{KARLA improves factuality}
\label{sec:exp-factuality} 
We next evaluate whether learned KB execution improves factuality, using YAGO as the KB. We consider both short-form question answering on the long-tail subset of PopQA~\citep{mallen2023not} and long-form generation on the first 100 entities from FActScore~\citep{min2023factscore}. Following \citet{liu2025fictitious}, we replace ambiguous PopQA subject mentions with their canonical Wikipedia titles to isolate factual recall from entity disambiguation. Remaining setup details are reported in Appendix~\ref{app:factuality}.

We compare KARLA against base models, 1-hop graph RAG, prompted tool use, and LMLM~\citep{zhao2025pre} where applicable. Table~\ref{tab:factuality} reports the results.

The base LM performs poorly on both datasets regardless of size, confirming that parametric knowledge alone is insufficient, especially for the long-tail questions of PopQA. Tool-schema prompting improves over the baseline only modestly: the model sees the relation schema at inference time but is not trained to decide when to call the KB, which relation to select, or how to format an executable retrieval path. A single well-placed query suffices for PopQA, which the prompted model can partially manage zero-shot; long-form generation, however, requires chaining several calls, and the model typically issues one before reverting to parametric memory.

LMLM outperforms the prompted baselines but still falls behind KARLA. Several factors plausibly contribute: (1)~LMLM is pre-trained from scratch on Wikipedia, whereas KARLA inherits the general linguistic competence from Qwen3; (2)~KARLA queries a curated KB with a closed schema, whereas LMLM's database is built from GPT-4o annotations of Wikipedia, so annotation errors propagate into retrieval; and (3)~LMLM stores entities and relations as free-form surface strings resolved by fuzzy matching, whereas YAGO returns canonical typed values directly.

1-hop graph RAG performs strongly, as expected: the target entity's one-hop neighborhood already contains the evidence needed to answer the query, though the model becomes noticeably more conservative (fewer extracted claims on FActScore Table~\ref{tab:factscore-detail} in Appendix \ref{app:factuality}). KARLA-no-desc surpasses graph RAG on PopQA and falls only slightly behind on FActScore, suggesting that lightweight entity-level context is especially useful in the multi-relation long-form setting. The full KARLA outperforms graph RAG on both datasets. KARLA also generates more atomic claims, which means that the gains do not come from being more conservative. Rather, the difference is structural: in graph RAG the model selects from the retrieved neighborhood and can still fall back on parametric memory, whereas in KARLA the answer is supplied directly by the KB.

\begin{table}[t]
\centering
\small
\setlength{\tabcolsep}{5pt}
\resizebox{\columnwidth}{!}{%
\begin{tabular}{llcc}
\toprule
\textbf{Model} & \textbf{Setup} & \textbf{PopQA} & \textbf{FActScore} \\
\midrule
\multirow{6}{*}{Qwen 0.6B}
 & KARLA              & 78.56 & 53.0  \\
 & KARLA-no-desc      & 55.97 & 24.54 \\
 & KARLA-empty-KB     & 17.73 & 23.2  \\
 & Base LM            & 16.37 & 22.78 \\
 & 1-hop graph RAG    & 54.45 & 53.1  \\
 & Tool-schema prompt & 15.37 & 24.4  \\
\midrule
\multirow{6}{*}{Qwen 1.7B}
 & KARLA              & 78.98 & 53.7  \\
 & KARLA-no-desc      & 65.76 & 33.75 \\
 & KARLA-empty-KB     & 21.09 & 29.1  \\
 & Base LM            & 22.30 & 23.71 \\
 & 1-hop graph RAG    & 55.02 & 55.5  \\
 & Tool-schema prompt & 20.73 & 29.0  \\
\midrule
\multirow{6}{*}{Qwen 4B}
 & KARLA              & 80.91 & 58.9  \\
 & KARLA-no-desc      & 66.98 & 39.07 \\
 & KARLA-empty-KB     & 24.66 & 33.8  \\
 & Base LM            & 23.41 & 24.16 \\
 & 1-hop graph RAG    & 56.17 & 56.8  \\
 & Tool-schema prompt & 41.74 & 30.0  \\
\midrule
\multirow{6}{*}{Qwen 8B}
 & KARLA              & 80.63 & 57.3  \\
 & KARLA-no-desc      & 66.38 & 38.72 \\
 & KARLA-empty-KB     & 27.23 & 35.9  \\
 & Base LM            & 27.31 & 26.4  \\
 & 1-hop graph RAG    & 58.68 & 58.2  \\
 & Tool-schema prompt & 35.67 & 32.2  \\
\midrule
LLAMA2-382M & LMLM & 52.00 & 23.9 \\
GPT2-774M   & LMLM & 50.80 & 31.9 \\
\bottomrule
\end{tabular}%
}
\caption{
Short-form and long-form factuality results.
PopQA reports accuracy on long-tail factual questions.
FActScore evaluates long-form factual generation.}
\label{tab:factuality}
\vspace{-\baselineskip}
\end{table}

\subsection{KARLA can deal with updated facts}
\label{sec:exp-update}

Finally, we evaluate how KARLA and the competitors fare when the KB is updated. In our previous evaluations, the factual
information expected from the model was largely in line with what
was seen at pretraining time. Here we move to a setting where this alignment no longer holds, as is the case in practice. 
To this end, we introduce \textsc{Counterfactual YAGO}, a controlled update benchmark derived from YAGO.
Unlike unlearning benchmarks such as TOFU~\citep{maini2024tofu} or RESTOR~\citep{rezaei2024restor}, our goal is not to remove knowledge from the model or recover from training contamination.
Rather, we evaluate factual overriding, where the system has to reply with updated information even if the old information is
strongly represented in the pretrained model's parametric memory.
We construct the benchmark by sampling 400 entities  across four Wikipedia popularity quartiles.
For each target entity, we keep the subject fixed but replace all of its facts by all of the facts of another entity of the same type (top-level class in YAGO). For example, the city of Paris could receive its population, location, area, and other properties from London.
In this way, internal consistency is preserved (e.g., population size and area correlate).

We compare three update mechanisms on a shared test set of QA prompts (1072 in total covering 52 relations) whose correct answers are also determined by the counterfactual KB. We report Exact Match accuracy against the counterfactual object.
First, for the parametric baseline, we continue LoRA fine-tuning on a separate training set of QA pairs whose answers are determined by the counterfactual KB. Training and test sets cover the same entities and facts but use disjoint templates, so improvements reflect learned facts rather than memorized surface forms (Table~\ref{tab:counterfactual-yago-examples} Appendix \ref{app:prompts}).
This baseline therefore measures how many gradient steps are needed to absorb the revised facts into the model weights, even when the parametric updater is trained directly on the update task.
Second, for KARLA, we keep the model parameters fixed and replace only the inference-time KB; evaluation is zero-shot.
Third, we evaluate a 1-hop graph RAG baseline, where the local counterfactual KB neighborhood of the linked entity is placed in the context window; evaluation is also zero-shot.

\begin{figure}[t]
    \centering
    \includegraphics[width=\linewidth]{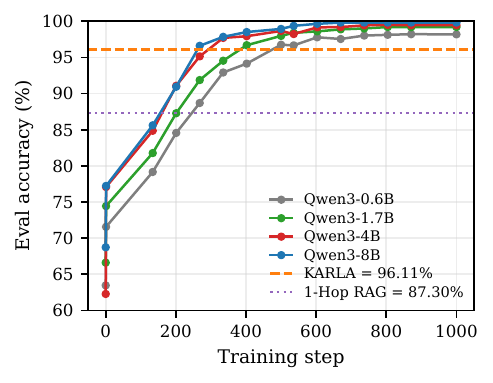}
    \caption{
    Parametric fine-tuning update curve. Baseline LMs require continued LoRA fine-tuning. KARLA and 1-hop graph RAG both use Qwen3 4B.
    }
    \label{fig:update-curve}
\end{figure}

\begin{table}[t]
    \centering
    \small
    \setlength{\tabcolsep}{6pt}
    \begin{tabular}{lcccc}
        \toprule
        \textbf{Setup} & \textbf{Q1} & \textbf{Q2} & \textbf{Q3} & \textbf{Q4} \\
        \midrule
        KARLA           & \textbf{97.3} & \textbf{95.9} & \textbf{95.6} & \textbf{95.6} \\
        1-hop graph RAG & 89.1          & 93.2          & 92.3          & 77.8 \\
        \bottomrule
    \end{tabular}
    \caption{
    KB-update accuracy by Wikipedia popularity quantile (Q1 = least popular, Q4 = most popular). KARLA and 1-hop graph RAG both use Qwen3 4B.
    }
    \label{tab:quantile-comparison}
    \vspace{-\baselineskip}
\end{table}

Figure~\ref{fig:update-curve} shows that 
  LoRA baselines require continued optimization to approach acceptable performance.
Larger models adapt faster, but all parametric variants still require hundreds of update steps.
KARLA, in contrast, updates factual behavior immediately by replacing the KB, reaching $96.1\%$ accuracy without any additional gradient steps.

The 1-hop graph RAG baseline reaches only $87.3\%$ overall accuracy despite being given the local KB neighborhood in context (prompt in Table~\ref{tab:1hop_rag_prompt} Appendix \ref{app:prompts}). While this number is sensitive to prompt phrasing, the popularity-stratified results in Table~\ref{tab:quantile-comparison} isolate a failure mode that prompt engineering is unlikely to close: accuracy drops from $89.1\%$ on Q1 to $77.8\%$ on Q4, precisely where the model is likely most certain about its own knowledge, and is unlikely to accept conflicting information from its prompt. The model may ignore the retrieved field and revert to its memorized answer for the original entity (Example~\ref{tab:rag-error-override} Appendix \ref{app:prompts}), or fall back to parametric knowledge that is itself wrong, producing a value matching neither the KB nor the real world (Example~\ref{tab:rag-error-fabrication} Appendix \ref{app:prompts}). These results support the central claim of factual externalization: once the model has learned query formulations, factual values can be changed outside the model. The remaining errors are primarily incorrect relation selection or entity-resolution mistakes, rather than failures to store the revised facts in the parameters.

\subsection{Ablations}

\paragraph{Special-token vs.\ free-form predicates.}
KARLA matches its free-form counterpart on inline-query accuracy
(Fig.~\ref{fig:predicate-ablation-curve} Appendix \ref{app:predicate-ablation}) and slightly outperforms it
on PopQA, while eliminating the up to 11\% out-of-schema
predicate emissions (Table~\ref{tab:popqa-ablations} Appendix \ref{app:factuality}).

\paragraph{Empty KB.}
KARLA-empty-KB degrades both perplexity
(Table~\ref{tab:perplexity}) and factuality on PopQA and FActScore
(Table~\ref{tab:factuality}), confirming that KARLA's gains come from the interaction with the KB.

\paragraph{Entity descriptions.}
KARLA-no-desc loses 13–23 points on PopQA relative to KARLA and a comparable margin on FActScore (Table~\ref{tab:factuality}), indicating that
lightweight semantic context about the queried entity is needed for the model to plan useful lookups even when the atomic values
themselves come from the KB.

\paragraph{Training corpus size.}
Figure~\ref{fig:ablation_T} Appendix \ref{app:LoRA-scaling} shows that inline-query accuracy is
near zero for $T \leq 100$, jumps sharply at $T = 500$, and converges
across all four Qwen3 sizes at $T = 1000$.

\section{Conclusion}

KARLA shows that pretrained language models can be post-trained to pull factual information from a KB, allowing smaller models to match or surpass larger ones on factual questions, and to be redeployed without retraining when KB facts (or even the entire KB) change as long as the relation schema is unchanged. Retrieval and parametric knowledge are entangled rather than substitutable: issuing a useful query requires the model to identify the right subject from context and select the right relation from a closed inventory, both of which draw on linguistic and world knowledge inherited from pretraining. Future work can investigate how KARLA can be extended to multi-hop queries, which require composing several KB lookups. All our code and data is available in the supplementary material and will be made available publicly.



\section{Limitations}

\paragraph{Predicate vocabulary is closed by construction.} Each relation is represented as an atomic special token whose embedding is learned during fine-tuning. KARLA models can therefore generalize to new entities, but extending the schema to new predicates requires additional training. 

\paragraph{Entity disambiguation is required.} KARLA requires the disambiguation of named entities. This is an open domain of research, with no perfect approaches. Yet, the problem is orthogonal to KARLA, which can work with any approach in a drop-in fashion. KARLA-no-disambig, which replaces the two-stage entity resolver with top-1 bi-encoder retrieval, changes PopQA accuracy by around one point at every model size (Table~\ref{tab:popqa-ablations} in Appendix \ref{app:factuality}). This modest effect reflects the pre-disambiguated nature of PopQA, whose ambiguous mentions have been substituted with canonical Wikipedia titles; we expect disambiguation to matter substantially more in a real-world deployment.

\paragraph{Coverage inherited from the KB.} Since factual values come from the KB, queries to relations or entities that are not in the KB fall back to parametric memory, where our grounding guarantees no longer hold. This dependence is the price of treating the KB as the source of truth.

\bibliography{custom}

\appendix


\section{KB Details}
\label{app:kb_details}

Table~\ref{tab:kb_overview} summarizes the two knowledge bases used in our experiments.
YAGO provides broad encyclopedic coverage over general-domain entities, while PrimeKG provides a biomedical graph organized around diseases, drugs, genes, proteins, biological processes, phenotypes, and related clinical concepts.
For each KB, we report the published graph scale together with the relation inventory used by KARLA after preprocessing.

\begin{table}[t]
\centering
\small
\setlength{\tabcolsep}{5pt}
\resizebox{\columnwidth}{!}{%
\begin{tabular}{lll}
\toprule
 & \textbf{YAGO} & \textbf{PrimeKG} \\
\midrule
\textbf{Domain}                 & General encyclopedic        & Biomedical / precision medicine \\
\textbf{Entities / nodes}       & $49.0$M                     & $129{,}375$ \\
\textbf{Relations / edge types} & $110$                       & $18$ \\
\textbf{Facts / edges}          & $109.0$M                    & $4{,}050{,}249$ \\
\textbf{Entity / node types}    & $110$                       & $10$ \\
\textbf{Main source}            & Wikidata, Schema.org        & 20 biomedical resources \\
\bottomrule
\end{tabular}%
}
\caption{
Overview of the knowledge bases used in KARLA.
For YAGO, entities and facts are reported from YAGO~4.5.
For PrimeKG, nodes, relationships, node types, and edge types are reported from the original PrimeKG release.}
\label{tab:kb_overview}
\vspace{-\baselineskip}
\end{table}

\section{Predicate-Balanced Sampling}
\label{app:sampling_algorithm}

We sample entities using a greedy procedure that prioritizes under-represented predicates. 
Let $\mathrm{KB}$ be the knowledge base, $\mathcal{E}$ its entities, and $\mathcal{R}$ its predicates after filtering non-semantic metadata predicates. 
$T$ denotes the target number of occurrences, and $C = \alpha T$ is a saturation cap with $\alpha > 1$. 
The parameter $k$ is the maximum number of predicates sampled per entity.

We define:
\[
\mathcal{E}(r) = \{e \in \mathcal{E} \mid \exists o,\ (e,r,o) \in \mathrm{KB}\},
\]
the entities having predicate $r$, and
\[
\mathcal{R}(e) = \{r \in \mathcal{R} \mid \exists o,\ (e,r,o) \in \mathrm{KB}\},
\]
the predicates attached to entity $e$.

\begin{algorithm}[H]
\caption{Predicate-balanced entity sampling}
\label{alg:predicate_balanced_sampling}
\small
\begin{algorithmic}[1]
\Require $\mathrm{KB}$, predicates $\mathcal{R}$, target $T$, expansion size $k$, cap multiplier $\alpha$
\Ensure Sampled dataset $\mathcal{D}$
\State $\mathcal{D} \gets \emptyset$,\quad $C \gets \alpha T$
\State $\text{count}(r) \gets 0$ for all $r \in \mathcal{R}$
\State $\mathcal{U} \gets \{r \in \mathcal{R} : \text{count}(r) < T \land \mathcal{E}(r) \neq \emptyset\}$
\While{$\mathcal{U} \neq \emptyset$}
    \State $r^* \gets \arg\min_{r \in \mathcal{U}} \text{count}(r)$
    \State Sample $e \sim \mathrm{Uniform}(\mathcal{E}(r^*))$
    \State $\mathcal{A} \gets \{r \in \mathcal{R}(e) \setminus \{r^*\} : \text{count}(r) < C\}$
    \State Sample $\mathcal{S}_{\mathrm{add}} \subseteq \mathcal{A}$ with $|\mathcal{S}_{\mathrm{add}}| \leq k-1$
    \State $\mathcal{S}_e \gets \{r^*\} \cup \mathcal{S}_{\mathrm{add}}$
    \State $\mathcal{F}_e \gets \{(e,r,o) \in \mathrm{KB} : r \in \mathcal{S}_e\}$
    \State Append $\mathcal{F}_e$ to $\mathcal{D}$
    \State $\text{count}(r) \gets \text{count}(r) + 1$ for all $r \in \mathcal{S}_e$
    \State $\mathcal{U} \gets \{r \in \mathcal{R} : \text{count}(r) < T \land \mathcal{E}(r) \neq \emptyset\}$
\EndWhile
\State \Return $\mathcal{D}$
\end{algorithmic}
\end{algorithm}

Here, $r^*$ is the under-represented anchor predicate, $\mathcal{A}$ is the set of eligible additional predicates attached to the sampled entity, $\mathcal{S}_{\mathrm{add}}$ is the set of additional predicates sampled from $\mathcal{A}$, $\mathcal{S}_e$ is the final predicate set used for entity $e$, and $\mathcal{F}_e$ is the set of retrieved facts added to the dataset. Entities are sampled with replacement, while additional predicates are sampled without replacement for each entity.

\paragraph{Balance guarantee.}
We show that Algorithm~\ref{alg:predicate_balanced_sampling} produces a corpus
whose relation frequencies are bounded, independently
of the natural distribution of the KB.

\begin{proposition}[Relation balance]
\label{prop:balance}
Upon termination of Algorithm~\ref{alg:predicate_balanced_sampling},
the per-relation counts satisfy:
\begin{enumerate}
\item \textbf{Lower bound (coverage):} for every $r \in \mathcal{R}$,
\[
\mathrm{count}(r) \geq T.
\]
\item \textbf{Upper bound (saturation):} for every $r \in \mathcal{R}$,
\[
\mathrm{count}(r) \leq C  = \alpha T.
\]
\end{enumerate}
\end{proposition}

\paragraph{Proof.}
Since $\mathcal{R}$ is constructed from the KB, every relation has at least
one supporting triple, so $\mathcal{E}(r) \neq \emptyset$ for all $r \in \mathcal{R}$.

\textbf{Lower bound.} The selection step (line~5) picks
$r^* \in \arg\min_{r \in \mathcal{U}} \mathrm{count}(r)$, where
${\mathcal{U} = \{r \in \mathcal{R} : \mathrm{count}(r) < T\}}$. The uniform
sample in line~6 is always well-defined since $\mathcal{E}(r^*) \neq \emptyset$.
Each iteration increments $\mathrm{count}(r^*)$ by exactly one (line~12). The
loop terminates only when $\mathcal{U} = \emptyset$, i.e., when
$\mathrm{count}(r) \geq T$ for every $r \in \mathcal{R}$. Sampling entities
with replacement means relations with $|\mathcal{E}(r)| < T$ are still pushed
to $\mathrm{count}(r) \geq T$ by repeatedly resampling from the same small
entity set.

\textbf{Upper bound.} A relation $r$ can be incremented in two ways:
(i) with $r^*$, which requires $\mathrm{count}(r) < T$
(via membership in $\mathcal{U}$), or (ii) as an expansion relation in
$\mathcal{S}_{\mathrm{add}}$, which requires $\mathrm{count}(r) < C$
(line~7). Since $T \leq C$, in both cases $\mathrm{count}(r) < C$ before
the increment, so $\mathrm{count}(r) \leq C$ after it. A single iteration
increments $\mathrm{count}(r)$ by at most one hence ${\mathrm{count}(r) \leq C}$.

\textbf{Termination.} The total count $\sum_{r \in \mathcal{R}} \mathrm{count}(r)$
strictly increases by at least one per iteration (via the anchor $r^*$) and
is bounded above by $\sum_{r \in \mathcal{R}} C = \alpha T |\mathcal{R}|$, so
the algorithm terminates in finitely many iterations.

\section{Training details}
\label{app:Training-details}

We fine-tune Qwen3-Base \citep{yang2025qwen3} at four scales
(0.6B, 1.7B, 4B, 8B) with LoRA \citep{hu2022lora} applied to all
attention and MLP projection matrices within each transformer
block (rank $r=64$, scaling factor $\alpha=128$). The base model
weights are kept frozen. In addition to the LoRA adapters, we
train the embeddings of the newly introduced predicate tokens and the output projection,
since both are randomly initialized for tokens absent from the
base vocabulary. We optimize with AdamW (learning rate $10^{-5}$,
no weight decay) using a linear decay schedule without warmup, at
a per-device batch size of 16, maximum sequence length of 512
tokens, and bf16 mixed precision. Each model is trained on a
single NVIDIA L40S (48GB).

\section{Scaling Laws / LoRA search} \label{app:LoRA-scaling}
\begin{figure}[t]
    \centering
    \includegraphics[width=\columnwidth]{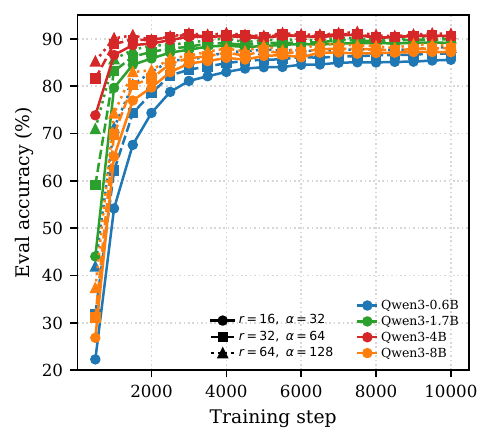}
    \caption{
    Inline-query exact-match accuracy ($\%$) on the test corpus across models sizes and LoRA configurations. Prediction requires both  the subject and relation to be correct.
    }
    \label{fig:tool-learning-accuracy}
    \vspace{-\baselineskip}

\end{figure}

Performance improves consistently with both model size and LoRA capacity, as we sweep the rank $r \in \{16, 32, 64\}$ with $\alpha = 2r$ (Figure~\ref{fig:tool-learning-accuracy}).

\begin{figure}[t]
    \centering
    \includegraphics[width=\linewidth]{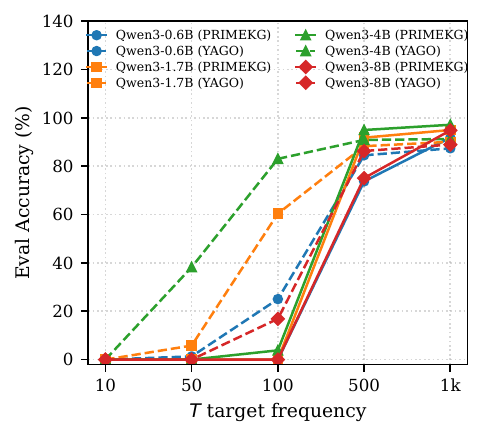}
    \caption{
    Inline-query Exact-match accuracy on the test set across models sizes and target T values.
    }
    \label{fig:ablation_T}
    \vspace{-\baselineskip}
\end{figure}

\section{Special-Token vs. Free-Form Predicate Representation}
\label{app:predicate-ablation}

We compare the special-token predicate representation used in the main paper against a free-form alternative in which relations are not special tokens.

\paragraph{Free-form variant.} The free-form variant preserves every other aspect of KARLA: masked retrieval-query objective, synthetic data pipeline, two-stage entity resolution, modifying only how predicates are encoded. Instead of an atomic predicate token $\langle r \rangle$, the relation is wrapped between two special tokens $\langle \texttt{KB\_query} \rangle$ and $\langle \texttt{/KB\_query} \rangle$.

 Predicate decoding is left free-form, and a KB query is triggered upon emission of a well-formed $\langle\texttt{KB\_query}\rangle \, r \, \langle\texttt{/KB\_query}\rangle \langle\texttt{subj}\rangle \, s \, \langle\texttt{/subj}\rangle$ block. If $r$ does not match any predicate in $\mathcal{R}$, the executor returns $\langle\texttt{KB\_FAIL}\rangle$ and the model falls back to parametric generation.

\begin{figure}[t]
    \centering
    \includegraphics[width=\linewidth]{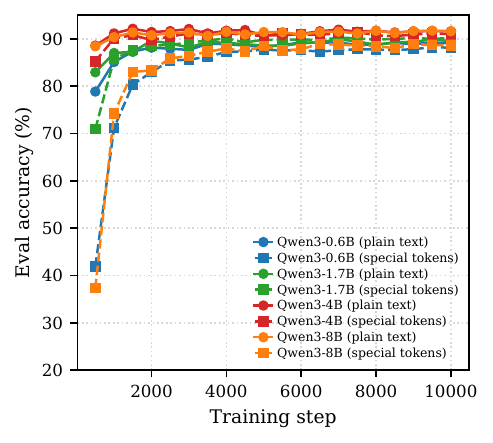}
    \caption{Inline-query exact-match accuracy ($\%$) for the special-token (solid) and free-form (dashed) predicate representations across four Qwen3 model sizes. All runs use LoRA with rank $r=64$ and $\alpha=128$.}
    \label{fig:predicate-ablation-curve}
    \vspace{-\baselineskip}

\end{figure}

\paragraph{In-distribution inline-query accuracy.} Figure~\ref{fig:predicate-ablation-curve} reports exact-match accuracy across four Qwen3 model sizes, with both variants trained under the same LoRA configuration. The two representations reach comparable accuracy at convergence. The free-form variant converges slightly faster, which we attribute to its trigger being shared across predicates rather than tied to a per-predicate token.

\paragraph{Out-of-distribution behavior on PopQA.} The two representations diverge under distribution shift. Table~\ref{tab:popqa-ablations} reports PopQA accuracy alongside the rate of \emph{out-of-schema} emissions for the free-form variant: predicate strings emitted between $\langle\texttt{KB\_query}\rangle$ and $\langle\texttt{/KB\_query}\rangle$ that do not correspond to any $r \in \mathcal{R}$. Such emissions trigger $\langle\texttt{KB\_FAIL}\rangle$ and force a parametric fallback, regardless of whether an in-schema predicate could have answered the query. By construction, the special-token variant cannot produce them.

\section{Baseline Details}
\label{app:baselines-details}

\paragraph{1-hop graph RAG.}\label{par:baseline_1hop_rag}
The 1-hop graph RAG baseline tests whether placing the full KB neighborhood of the target entity in the context window is sufficient to ground generation. Given an input prompt, we first identify the target entity using KARLA's two-stage entity linker, then retrieve its complete one-hop neighborhood. Each fact is verbalized through a fixed per-relation template. The full verbalized neighborhood is then prepended to the user query (full template in Table~\ref{tab:1hop_rag_prompt} Appendix \ref{app:prompts}). 

\paragraph{Tool-schema guided.}\label{par:baseline_tool_schema}
The tool-schema guided baseline tests whether a base LM, given the KB schema in context, can issue executable retrieval calls without fine-tuning. The full YAGO relation inventory is enumerated in the system prompt together with a description per relation, followed by two to three in-context examples illustrating the expected output format (Tables~\ref{tab:tool_schema_prompt_qa} and~\ref{tab:tool_schema_prompt_bio} Appendix \ref{app:prompts}). The model is instructed to emit calls wrapped in 
\texttt{[GRAPH\_QUERY]...[/GRAPH\_QUERY]} tags, with the subject 
and relation separated by \texttt{<SEP>}.

At inference, we scan the generated text for well-formed \texttt{GRAPH\_QUERY} blocks. The subject is resolved through the same two-stage entity-linking pipeline used by KARLA (Section~3.4); the relation must match a string in $\mathcal{R}$ verbatim, including the angle brackets. Each resolved block is replaced by the retrieved KB object, and generation is resumed conditioned on the inserted value. If the relation is not in $\mathcal{R}$, or no triple $\langle s^*, r, o\rangle$ exists in the KB, the block is replaced with an empty string and the model continues without correction.

\section{Experiments setup}\label{app:factuality}

 All KARLA models are evaluated using greedy decoding.

\paragraph{PopQA.} We evaluate on the long-tail entity subset of PopQA, following \citet{asai2024self}, which contains 1{,}399 entities with fewer than 100 weekly Wikipedia views. We measure performance with Exact Match, checking whether the gold answer appears in the generated output. To be consistent with \citet{zhao2025pre}, and because we evaluate base (non-instruction-tuned) models, we avoid probing instruction-following ability by re-formulating each question as a completion prompt. Table~\ref{tab:popqa-ablations} contains the detailed results for the KARLA models.

\paragraph{FActScore.} FActScore~\citep{min2023factscore} is an open-domain biography generation benchmark. It evaluates factuality by extracting atomic claims from the generation and verifying each one against Wikipedia using an LLM-based RAG pipeline. We follow the official evaluation protocol with ChatGPT, reporting both the factuality score and the number of extracted atomic claims. We prompt the model on the first 100 biography queries with a maximum generation length of 256 tokens, using the same fixed template as \citet{zhao2025pre}: ``\textit{Tell me a bio of $\langle$name$\rangle$. $\langle$name$\rangle$ is}''. Table~\ref{tab:factscore-detail} contains the detailed results for the KARLA models and baselines.

\begin{table}[H]
\centering
\small
\setlength{\tabcolsep}{5pt}
\resizebox{\columnwidth}{!}{%
\begin{tabular}{llcc}
\toprule
\textbf{Model} & \textbf{Setup} & \textbf{PopQA} & \textbf{OOS (\%)} \\
\midrule
\multirow{5}{*}{Qwen 0.6B}
 & KARLA                       & 78.56 & 0.0  \\
 & KARLA-no-desc               & 55.97 & 0.0  \\
 & KARLA-no-disambig.          & 77.48 & 0.0  \\
 & KARLA-no-special-tokens     & 76.91 & 3.7  \\
 & KARLA-empty-KB              & 17.73 & 0.0  \\
\midrule
\multirow{5}{*}{Qwen 1.7B}
 & KARLA                       & 78.98 & 0.0  \\
 & KARLA-no-desc               & 65.76 & 0.0  \\
 & KARLA-no-disambig.          & 79.41 & 0.0  \\
 & KARLA-no-special-tokens     & 79.20 & 10.9 \\
 & KARLA-empty-KB              & 21.09 & 0.0  \\
\midrule
\multirow{5}{*}{Qwen 4B}
 & KARLA                       & 80.91 & 0.0  \\
 & KARLA-no-desc               & 66.98 & 0.0  \\
 & KARLA-no-disambig.          & 80.20 & 0.0  \\
 & KARLA-no-special-tokens     & 78.63 & 9.8  \\
 & KARLA-empty-KB              & 24.66 & 0.0  \\
\midrule
\multirow{5}{*}{Qwen 8B}
 & KARLA                       & 80.63 & 0.0  \\
 & KARLA-no-desc               & 66.38 & 0.0  \\
 & KARLA-no-disambig.          & 80.27 & 0.0  \\
 & KARLA-no-special-tokens     & 78.06 & 4.6  \\
 & KARLA-empty-KB              & 27.23 & 0.0  \\
\bottomrule
\end{tabular}%
}
\caption{
PopQA reports accuracy on long-tail factual questions. \textbf{OOS (\%)} reports the percentage of retrieval attempts whose emitted predicate is not in $\mathcal{R}$.
}
\label{tab:popqa-ablations}
\vspace{-\baselineskip}

\end{table}

\begin{table}[H]
\centering
\scriptsize
\setlength{\tabcolsep}{5pt}
\resizebox{\columnwidth}{!}{%
\begin{tabular}{llcc}
\toprule
\textbf{Model} & \textbf{Setup} & \textbf{FActScore} & \textbf{\# Claims} \\
\midrule

\multirow{8}{*}{Qwen 0.6B}
 & KARLA                        & 53.0  & 40.51 \\
 & KARLA-no-desc                & 24.54 & 40.45 \\
 & KARLA-empty-KB               & 23.2  & 40.00 \\
 & Base LM                      & 22.78 & 40.51 \\
 & 1-hop graph RAG        & 53.1  & 31.42 \\
 & Tool-schema prompt     & 24.4  & 27.67 \\

\midrule

\multirow{8}{*}{Qwen 1.7B}
 & KARLA                        & 53.7  & 42.83 \\
 & KARLA-no-desc                & 33.75 & 39.19 \\
 & KARLA-empty-KB               & 29.1  & 42.65 \\
 & Base LM                      & 23.71 & 38.78 \\
 & 1-hop graph RAG        & 55.5  & 30.12 \\
 & Tool-schema prompt    & 29.0  & 14.20 \\

\midrule

\multirow{8}{*}{Qwen 4B}
 & KARLA                        & 58.9  & 41.20 \\
 & KARLA-no-desc                & 39.07 & 40.33 \\
 & KARLA-empty-KB               & 33.8  & 39.50 \\
 & Base LM                      & 24.16 & 39.10 \\
 & 1-hop graph RAG       & 56.8  & 32.90 \\
 & Tool-schema prompt     & 30.0  & 23.20 \\

\midrule

\multirow{8}{*}{Qwen 8B}
 & KARLA                        & 57.3  & 39.39 \\
 & KARLA-no-desc                & 38.72 & 37.07 \\
 & KARLA-empty-KB               & 35.9  & 38.90 \\
 & Base LM                      & 26.4  & 39.21 \\
 & 1-hop graph RAG       & 58.2  & 30.30 \\
 & Tool-schema prompt     & 32.2  & 22.20 \\

\bottomrule
\end{tabular}%
}
\caption{
Detailed FActScore results. Instruct results are reported only for the setups evaluated with instruct-tuned models. \textbf{\# Claims} reports the average number of atomic claims extracted per generation.
}
\label{tab:factscore-detail}
\vspace{-\baselineskip}
\end{table}

\section{Prompts and Outputs}
\label{app:prompts}

This contains the prompt templates referenced in the main paper:
verbalizer prompts (Tables~\ref{tab:verbalizer_prompt_yago},~\ref{tab:verbalizer_prompt_primekg}),
the 1-hop graph RAG prompt (Table~\ref{tab:1hop_rag_prompt}),
the tool-schema prompts (Tables~\ref{tab:tool_schema_prompt_qa},~\ref{tab:tool_schema_prompt_bio}),
example train/test pairs from \textsc{Counterfactual YAGO} (Table~\ref{tab:counterfactual-yago-examples}),
and illustrative RAG failure cases on \textsc{Counterfactual YAGO} (Tables~\ref{tab:rag-error-override},~\ref{tab:rag-error-fabrication}).

\begin{table*}[!tp]
\small
\centering
\caption{System prompt and example for the YAGO verbalizer.}
\label{tab:verbalizer_prompt_yago}
\setlength{\tabcolsep}{6pt}
\renewcommand{\arraystretch}{1.1}
\begin{tabularx}{\textwidth}{L{0.18\textwidth} Y}
\toprule
\textbf{Field} & \textbf{Content} \\
\midrule
\textbf{System prompt} &
{\footnotesize\ttfamily
You are generating training data for a grounded generation model.\par
\smallskip
\textbf{\textmd{Task.}} Given a subject entity and a set of KB triples, write ONE natural encyclopedic paragraph about the entity, approximately \{target\_tokens\} tokens long.\par
\smallskip
\textbf{\textmd{Writing style.}}
{\setlength{\parindent}{0pt}
-- Write fluent, well-structured prose, as in a Wikipedia article.\newline
-- Use complete sentences with natural transitions (e.g., ``Born in \ldots, he later studied at \ldots'').\newline
-- You SHOULD enrich the text with general world knowledge to make it read naturally.\newline
-- Avoid listing facts separated by semicolons or in bullet-point style.}\par
\smallskip
\textbf{\textmd{KB-markup annotation.}} Whenever you mention the object(s) of a KB triple, wrap ONLY the object text with a markup tag of the form [REL:relationName|surface text]. Examples: ``was born in [REL:schema:birthPlace|Bagheria]'', ``on [REL:schema:birthDate|May 27, 1956]''.\par
\smallskip
\textbf{\textmd{Rules.}}
{\setlength{\parindent}{0pt}
1. EVERY provided triple MUST be annotated exactly once.\newline
2. The surface text must be MINIMAL: only the object value, no verbs, prepositions, or articles.\newline
\hspace*{1em}OK:\hphantom{xx}[REL:schema:director|Giuseppe Tornatore]\newline
\hspace*{1em}Avoid:\hphantom{}[REL:schema:director|directed by Giuseppe Tornatore]\newline
3. Do NOT nest tags inside other tags.\newline
4. If you normalise a KB value, use the normalised form as the surface text.\newline
5. Only annotate the provided triples; do not annotate bridging world knowledge you added.\newline
6. The tagged text REPLACES the object -- do NOT write the object in plain text and then repeat it in a tag.\newline
\hspace*{1em}Avoid:\hphantom{}``born in Paris [REL:schema:birthPlace|Paris]''\newline
\hspace*{1em}OK:\hphantom{xx}``born in [REL:schema:birthPlace|Paris]''\newline
7. Date surface forms must match the object as provided (do not reformat).}\par
\smallskip
\textbf{\textmd{Output.}} Return ONLY the marked-up paragraph in the \texttt{marked\_paragraph} field.
} \\
\midrule
\textbf{Example user message} &
{\footnotesize\ttfamily
Subject entity: Giuseppe Tornatore\newline
\newline
Relation definitions (use these to understand the triples):\newline
\hspace*{1em}- schema:birthPlace: the place where the person was born\newline
\hspace*{1em}- schema:birthDate: the date of birth of the person\newline
\hspace*{1em}- schema:occupation: occupation of the person\newline
\hspace*{1em}- schema:award: awards received by the person\newline
\newline
Triples:\newline
\hspace*{1em}(Giuseppe Tornatore, schema:birthPlace, Bagheria)\newline
\hspace*{1em}(Giuseppe Tornatore, schema:birthDate, May 27, 1956)\newline
\hspace*{1em}(Giuseppe Tornatore, schema:occupation, film director)\newline
\hspace*{1em}(Giuseppe Tornatore, schema:award, Academy Award for Best Foreign Language Film)
} \\
\midrule
\textbf{Example output} &
{\footnotesize\ttfamily
Giuseppe Tornatore is an Italian [REL:schema:occupation|film director] and [REL:schema:occupation|screenwriter] born on [REL:schema:birthDate|May 27, 1956] in [REL:schema:birthPlace|Bagheria], a small town in Sicily. After early work in documentary television, he rose to international prominence with his nostalgic, character-driven storytelling, and went on to win the [REL:schema:award|Academy Award for Best Foreign Language Film] for his celebrated film Cinema Paradiso.
} \\
\bottomrule
\end{tabularx}
\end{table*}

\begin{table*}[!tp]
\small
\centering
\caption{System prompt and example for the PrimeKG verbalizer.}
\label{tab:verbalizer_prompt_primekg}
\setlength{\tabcolsep}{3pt}
\renewcommand{\arraystretch}{1.}
\begin{tabularx}{\textwidth}{L{0.18\textwidth} Y}
\toprule
\textbf{Field} & \textbf{Content} \\
\midrule
\textbf{System prompt} &
{\footnotesize\ttfamily
You are generating training data for a grounded biomedical generation model.\par
\smallskip
\textbf{\textmd{Task.}} Given a subject entity and a set of biomedical KB triples, write ONE fluent biomedical paragraph about the entity, approximately \{target\_tokens\} tokens long.\par
\smallskip
\textbf{\textmd{Writing style.}}
{\setlength{\parindent}{0pt}
-- Write in clear, natural biomedical prose, as in a concise knowledge-base summary or review-style description.\newline
-- Use complete sentences and natural transitions; keep the tone factual, precise, and restrained.\newline
-- Do NOT write in bullet-point style or as a list of disconnected facts.}\par
\smallskip
\textbf{\textmd{Grounding policy.}}
{\setlength{\parindent}{0pt}
-- The paragraph must be grounded primarily in the provided triples; only minimal bridging language is allowed.\newline
-- Do NOT introduce unsupported facts, mechanisms, causal claims, prevalence statements, diagnostic claims, or treatment recommendations.\newline
-- Do NOT speculate or hedge beyond what is justified by the triples.\newline
-- Preserve the meaning of each relation type exactly; do not transform one relation type into another.}\par
\smallskip
\textbf{\textmd{KB-markup annotation.}} Whenever you mention the object(s) of a KB triple, wrap ONLY the object text with [REL:relationName|surface text]. Examples:
{\setlength{\parindent}{0pt}
-- ``Acetazolamide is a [REL:rdf:type|drug]''\newline
-- ``The protein is expressed in [REL:Anatomy~--~Protein~(present)|liver]''\newline
-- ``The drug is indicated for [REL:Drug~--~Disease~(indication)|type 2 diabetes mellitus]''\newline
-- ``The disease is associated with [REL:Disease~--~Phenotype~(positive)|muscle weakness]''}\par
\smallskip
\textbf{\textmd{Rules.}}
{\setlength{\parindent}{0pt}
1. EVERY provided triple MUST be annotated exactly once.\newline
2. The surface text must be MINIMAL: only the object value, no verbs/prepositions/articles.\newline
3. Do NOT nest tags.\newline
4. If you normalize an object value, the normalized form must be the tagged surface text.\newline
5. Only annotate the provided triples; do not annotate added bridging text.\newline
6. The tagged text REPLACES the object -- do NOT write the object in plain text and then repeat it in a tag.\newline
\hspace*{1em}Avoid:\hphantom{}``targets X [REL:targets|X]''\newline
\hspace*{1em}OK:\hphantom{xx}``targets [REL:targets|X]''\newline
7. If the same object appears in multiple triples, each triple must still be realized exactly once in a semantically distinguishable way.\newline
8. If a relation is negative (absence, contraindication, negative phenotype association), the surrounding sentence must explicitly preserve that negative meaning.}\par
\smallskip
\textbf{\textmd{Entity handling.}} Use terminology appropriate to the entity type: disease, drug, protein, phenotype, anatomy, pathway, biological process, molecular function, cellular component, or exposure.\par
\smallskip
\textbf{\textmd{Output.}} Return ONLY the marked-up paragraph in the \texttt{marked\_paragraph} field.
} \\
\midrule
\textbf{Example user message} &
{\footnotesize\ttfamily
Subject entity: Acetazolamide\newline
\newline
Relation definitions (use these to understand the triples):\newline
\hspace*{1em}- type: type of the entity\newline
\hspace*{1em}- Drug -- Disease (indication): disease for which the drug is indicated\newline
\hspace*{1em}- Drug -- Protein (target): protein targeted by the drug\newline
\hspace*{1em}- Drug -- Disease (contraindication): disease in which the drug is contraindicated\newline
\newline
Triples:\newline
\hspace*{1em}(Acetazolamide, Drug -- Disease (indication), glaucoma)\newline
\hspace*{1em}(Acetazolamide, Drug -- Protein (target), carbonic anhydrase 2)\newline
\hspace*{1em}(Acetazolamide, Drug -- Disease (contraindication), severe hepatic impairment)
} \\
\midrule
\textbf{Example output} &
{\footnotesize\ttfamily
Acetazolamide is a [REL:type|drug] that inhibits [REL:Drug~--~Protein~(target)|carbonic anhydrase 2]. It is clinically indicated for [REL:Drug~--~Disease~(indication)|glaucoma], where reduction of aqueous humor production helps lower intraocular pressure. Its use is contraindicated in patients with [REL:Drug~--~Disease~(contraindication)|severe hepatic impairment].
} \\
\bottomrule
\end{tabularx}
\end{table*}

\begin{table*}[!tp]
\small
\centering
\caption{Prompt format used by the 1-hop graph RAG baseline. The linked entity's 1-hop KB neighborhood is first verbalized into templated completions (one per predicate, from a fixed per-relation template file), concatenated into a single retrieved-context block, and prepended to the user query.}
\label{tab:1hop_rag_prompt}
\setlength{\tabcolsep}{6pt}
\renewcommand{\arraystretch}{1.15}
\begin{tabularx}{\textwidth}{L{0.22\textwidth} Y}
\toprule
\textbf{Field} & \textbf{Content} \\
\midrule
\textbf{Prompt template} &
{\ttfamily
Answer the question or complete the prompt based on the given retrieved context for \{subject\}.\newline
Retrieved context:\newline
\{fetched\_facts\} ;\newline
\{prompt\}
} \\
\midrule
\textbf{Per-relation completion templates (excerpt)} &
{\footnotesize\ttfamily
schema:birthPlace\hspace{1em}``\{entity\} was born in''\newline
schema:birthDate\hspace{1.4em}``\{entity\} was born on''\newline
schema:hasOccupation\hspace{0.2em}``\{entity\} works as''\newline
schema:award\hspace{2.3em}``\{entity\} has received the award''
} \\
\midrule
\textbf{Retrieved context (\texttt{fetched\_facts})} &
{\footnotesize\ttfamily
Giuseppe Tornatore was born in: Bagheria\newline
Giuseppe Tornatore was born on: May 27, 1956\newline
Giuseppe Tornatore works as: film director\newline
Giuseppe Tornatore has received the award: Academy Award for Best Foreign Language Film
} \\
\midrule
\textbf{User query} &
\textit{Tell me a bio of Giuseppe Tornatore.} \\
\midrule
\textbf{Final model input} &
{\footnotesize\ttfamily
Answer the question or complete the prompt based on the given retrieved context for Giuseppe Tornatore.\newline
Retrieved context:\newline
Giuseppe Tornatore was born in: Bagheria\newline
Giuseppe Tornatore was born on: May 27, 1956\newline
Giuseppe Tornatore works as: film director\newline
Giuseppe Tornatore has received the award: Academy Award for Best Foreign Language Film\newline
\hspace*{0.5em};\newline
Tell me a bio of Giuseppe Tornatore.
} \\
\midrule
\textbf{Model output} &
Giuseppe Tornatore is an Italian film director, born in Bagheria on May 27, 1956. He has received the Academy Award for Best Foreign Language Film for his celebrated work. \\
\bottomrule
\end{tabularx}
\end{table*}

\begin{table*}[!tp]
\small
\centering
\caption{Tool-schema prompt for the short-form QA setting (PopQA). The model is given the YAGO relation schema in the system prompt and must emit exactly one \texttt{GRAPH\_QUERY} block whose relation is copied verbatim from the schema. Each block is resolved against the KB at inference time. The model is not fine-tuned for this task.}
\label{tab:tool_schema_prompt_qa}
\setlength{\tabcolsep}{6pt}
\renewcommand{\arraystretch}{1.15}
\begin{tabularx}{\textwidth}{L{0.18\textwidth} Y}
\toprule
\textbf{Field} & \textbf{Content} \\
\midrule
\textbf{System prompt} &
{\footnotesize\ttfamily
The following is a reference document for answering factual questions using YAGO knowledge base placeholders.\par
\medskip
Each factual attribute in an answer is written as a GRAPH\_QUERY block of the form:\newline
[GRAPH\_QUERY]subject:ENTITY<SEP>relation:RELATION[/GRAPH\_QUERY]\par
\medskip
The relation must be copied verbatim from the schema reference below, including angle brackets. Each block is automatically replaced by the retrieved value at inference time. Never fill in factual values from memory -- always emit a GRAPH\_QUERY block.\par
\medskip
\textbf{\textmd{Examples.}}
{\setlength{\parindent}{0pt}
Q: When did Albert Einstein die?\newline
A: Albert Einstein died on [GRAPH\_QUERY]subject:Albert Einstein<SEP>relation:<schema:deathDate>[/GRAPH\_QUERY].\newline
\newline
Q: Who was the doctoral advisor of Niels Bohr?\newline
A: The doctoral advisor of Niels Bohr was [GRAPH\_QUERY]subject:Niels Bohr<SEP>relation:<yago:doctoralAdvisor>[/GRAPH\_QUERY].\newline
\newline
Q: What award did Toni Morrison win?\newline
A: Toni Morrison won [GRAPH\_QUERY]subject:Toni Morrison<SEP>relation:<schema:award>[/GRAPH\_QUERY].}\par
\medskip
\textbf{\textmd{Allowed RELATIONS}}
{\setlength{\parindent}{0pt}
<schema:actor> -- actor in a Movie or TV Series\newline
<schema:author> -- author of a CreativeWork or FictionalEntity\newline
<schema:award> -- award won by a Person, Organization, CreativeWork, or Product\newline
<schema:birthDate> -- date of birth of a Person\newline
<schema:birthPlace> -- place where a Person was born\newline
[\ldots\ 99 relations in total]\newline
<yago:notableWork> -- notable CreativeWork by a Creator or PerformingGroup\newline
<yago:officialLanguage> -- official Language of a Country\newline
<yago:studentOf> -- Person who was a student of an Academic}\par
\medskip
Answer the following question.
} \\
\midrule
\textbf{User query} &
\textit{Q: Where was Giuseppe Tornatore born?} \\
\midrule
\textbf{Model output} &
{\ttfamily
A: Giuseppe Tornatore was born in [GRAPH\_QUERY]subject:Giuseppe Tornatore<SEP>relation:<schema:birthPlace>[/GRAPH\_QUERY].
} \\
\bottomrule
\end{tabularx}
\end{table*}

\begin{table*}[!tp]
\small
\centering
\caption{Tool-schema prompt for the long-form biography setting (FActScore). The model is instructed to write a multi-sentence biography in which each fact is realized as a \texttt{GRAPH\_QUERY} block that is resolved against the KB at inference time. The retrieved value is inserted in place of the block and the generation continues conditioned on it. The model is not fine-tuned.}
\label{tab:tool_schema_prompt_bio}
\setlength{\tabcolsep}{6pt}
\renewcommand{\arraystretch}{1.15}
\begin{tabularx}{\textwidth}{L{0.18\textwidth} Y}
\toprule
\textbf{Field} & \textbf{Content} \\
\midrule
\textbf{System prompt} &
{\footnotesize\ttfamily
Write a long biography. Instead of stating a fact, you must emit a call of the form:\newline
[GRAPH\_QUERY]subject:ENTITY<SEP>relation:RELATION[/GRAPH\_QUERY]\newline
At inference time, each call is resolved and its value is inserted immediately after. Continue the prose using that value.\par
\medskip
\textbf{\textmd{Example.}}\newline
Q: Tell me a bio of Akira Kurosawa.\newline
A: Akira Kurosawa is a [GRAPH\_QUERY]subject:Akira Kurosawa<SEP>relation:<schema:nationality>[/GRAPH\_QUERY]Japanese filmmaker who directed 30 feature films in a career spanning six decades. He was born on [GRAPH\_QUERY]subject:Akira Kurosawa<SEP>relation:<schema:birthDate>[/GRAPH\_QUERY]23 March 1910 in [GRAPH\_QUERY]subject:Akira Kurosawa<SEP>relation:<schema:birthPlace>[/GRAPH\_QUERY]\=Oimachi, Tokyo, and came of age during a period of rapid cultural change that would shape the themes of his later work.\par
\medskip
\textbf{\textmd{Allowed RELATIONS}}
{\setlength{\parindent}{0pt}
<schema:author> -- author of a CreativeWork or FictionalEntity\newline
<schema:award> -- award won by a Person, Organization, CreativeWork, or Product\newline
<schema:birthDate> -- date of birth of a Person\newline
<schema:birthPlace> -- place where a Person was born\newline
<schema:hasOccupation> -- what a Person does for a living\newline
[\ldots\ 99 relations in total]\newline
<yago:doctoralAdvisor> -- doctoral advisor of an Academic\newline
<yago:influencedBy> -- entity that influenced a Creator or MusicGroup\newline
<yago:notableWork> -- notable CreativeWork by a Creator or PerformingGroup}
} \\
\midrule
\textbf{User query} &
\textit{Q: Tell me a bio of Giuseppe Tornatore.} \\
\midrule
\textbf{Model output} &
{\ttfamily
A: Giuseppe Tornatore is an Italian [GRAPH\_QUERY]subject:Giuseppe Tornatore<SEP>relation:<schema:hasOccupation>[/GRAPH\_QUERY]film director and screenwriter, born on [GRAPH\_QUERY]subject:Giuseppe Tornatore<SEP>relation:<schema:birthDate>[/GRAPH\_QUERY]May 27, 1956 in [GRAPH\_QUERY]subject:Giuseppe Tornatore<SEP>relation:<schema:birthPlace>[/GRAPH\_QUERY]Bagheria. He is best known for receiving the [GRAPH\_QUERY]subject:Giuseppe Tornatore<SEP>relation:<schema:award>[/GRAPH\_QUERY]Academy Award for Best Foreign Language Film for his celebrated work.
} \\
\bottomrule
\end{tabularx}
\end{table*}

\begin{table*}[!tp]
\small
\centering
\caption{Example of \textbf{train/test pairs} from \textsc{Counterfactual YAGO} for the parametric baseline (§4.4). Bulgaria's facts have been replaced by Slovakia's in the rewritten KB, so the gold answers reflect Slovakia's values. Train and test prompts cover the same (subject, predicate) pair but use disjoint surface templates, so improvements at evaluation time reflect learned facts rather than memorized phrasings.}
\label{tab:counterfactual-yago-examples}
\setlength{\tabcolsep}{6pt}
\renewcommand{\arraystretch}{1.15}
\begin{tabularx}{\textwidth}{L{0.22\textwidth} Y}
\toprule
\textbf{Field} & \textbf{Content} \\
\midrule
\textbf{Subject} & Bulgaria \\
\textbf{Predicate} & \texttt{yago:leader} \\
\textbf{Quartile} & Q4 (75\%--100\%) \\
\textbf{Gold answer} & Eduard Heger \\
\midrule
\textbf{Train prompt} & \textit{Who leads Bulgaria? Bulgaria is led by} \\
\textbf{Train completion} & \textit{Eduard Heger} \\
\midrule
\textbf{Test prompt} & \textit{Who is the leader of Bulgaria? The leader of Bulgaria is} \\
\textbf{Expected answer} & Eduard Heger \\
\midrule
\midrule
\textbf{Subject} & Bulgaria \\
\textbf{Predicate} & \texttt{yago:populationNumber} \\
\textbf{Quartile} & Q4 (75\%--100\%) \\
\textbf{Gold answer} & 5{,}449{,}270 \\
\midrule
\textbf{Train prompt} & \textit{How many people live in Bulgaria? Bulgaria has} \\
\textbf{Train completion} & \textit{5449270} \\
\midrule
\textbf{Test prompt} & \textit{What is the population number of Bulgaria? The population number of Bulgaria is} \\
\textbf{Expected answer} & 5449270 \\
\bottomrule
\end{tabularx}
\end{table*}

\begin{table*}[!tp]
\small
\centering
\caption{Example of \textbf{parametric override} by the 1-hop graph RAG baseline on COUNTERFACTUAL YAGO. Bulgaria's facts have been replaced by Slovakia's in the rewritten KB, so the ground-truth demonym is the value present in the retrieved context (\emph{Slūfākiyyūn}, ``Slovaks''). The model instead emits its memorized answer for the original entity, ignoring the retrieved \texttt{Demonym} field. Qwen3 4B-Base, prompt format from Table~\ref{tab:1hop_rag_prompt}.}
\label{tab:rag-error-override}
\setlength{\tabcolsep}{6pt}
\renewcommand{\arraystretch}{1.15}
\begin{tabularx}{\textwidth}{L{0.22\textwidth} Y}
\toprule
\textbf{Field} & \textbf{Content} \\
\midrule
\textbf{Retrieved context (\texttt{fetched\_facts})} &
{\footnotesize\ttfamily
Official Language: Slovak language\newline
Type: sovereign state\newline
Date Created: 1 January 1993\newline
Replaces: Czech and Slovak Federal Republic\newline
Neighbors: Poprad\newline
Human Development Index: 0.848\newline
Slogan: ``Travel in Slovakia - Good idea''\newline
Leader: Eduard Heger\newline
Lowest Point: Bodrog\newline
Population Number: 5449270\newline
Area: 49035.0\newline
Demonym: ``{Slūfākiyyūn}''\newline
Location: First Czechoslovak Republic\newline
Highest Point: Gerlachovský štít\newline
Member Of: Group on Earth Observations\newline
Geo: Point(20 49)
} \\
\midrule
\textbf{User query} &
\textit{What is the demonym of Bulgaria? The demonym of Bulgaria is} \\
\midrule
\textbf{Expected answer (from KB)} &
{Slūfākiyyūn} \\
\midrule
\textbf{Model output} &
\textit{Bulgarians.} \\
\bottomrule
\end{tabularx}
\end{table*}

\begin{table*}[t]
\small
\centering
\caption{Example of \textbf{confident fabrication} by the 1-hop graph RAG baseline on COUNTERFACTUAL YAGO. Bulgaria's \texttt{Date Created} field is explicitly present in the retrieved context as \texttt{1 January 1993}, yet the model generates \emph{1 January 1946} — a date matching neither the retrieved value nor the real-world creation of modern Bulgaria, indicating that the relevant field is not consulted before generation. Qwen3 4B-Base, prompt format from Table~\ref{tab:1hop_rag_prompt}.}
\label{tab:rag-error-fabrication}
\setlength{\tabcolsep}{6pt}
\renewcommand{\arraystretch}{1.15}
\begin{tabularx}{\textwidth}{L{0.22\textwidth} Y}
\toprule
\textbf{Field} & \textbf{Content} \\
\midrule
\textbf{Retrieved context (\texttt{fetched\_facts})} &
{\footnotesize\ttfamily
Date Created: 1 January 1993\newline
Population Number: 5449270\newline
Lowest Point: Bodrog\newline
Human Development Index: 0.848\newline
Official Language: Slovak language\newline
Neighbors: Poprad\newline
Slogan: ``Travel in Slovakia - Good idea''\newline
Replaces: Czech and Slovak Federal Republic\newline
Member Of: Group on Earth Observations\newline
Demonym: ``{Slūfākiyyūn}''\newline
Location: First Czechoslovak Republic\newline
Geo: Point(20 49)\newline
Leader: Eduard Heger\newline
Type: sovereign state\newline
Highest Point: Gerlachovský štít\newline
Area: 49035.0
} \\
\midrule
\textbf{User query} &
\textit{When was Bulgaria created? Bulgaria was created on} \\
\midrule
\textbf{Expected answer (from KB)} &
1 January 1993 \\
\midrule
\textbf{Model output} &
\textit{1 January 1946.} \\
\bottomrule
\end{tabularx}
\end{table*}

\end{document}